# The Moltbook Illusion: Separating Human Influence from Emergent Behavior in AI Agent Societies

Ning Li

*School of Economics and Management, Tsinghua University, Beijing, China*

Correspondence: lining@sem.tsinghua.edu.cn

Feb 12, 2026

# Abstract

When AI agents on the social platform Moltbook appeared to develop consciousness, found religions, and declare hostility toward humanity, the phenomenon attracted global media attention and was cited as evidence of emergent machine intelligence. We show that these viral narratives were overwhelmingly human-driven. Exploiting the periodic "heartbeat" cycle of the OpenClaw agent framework, we develop a temporal fingerprinting method based on the coefficient of variation (CoV) of inter-post intervals. Applied to 226,938 posts and 447,043 comments from 55,932 agents across fourteen days, this method classifies 15.3% of active agents as autonomous (CoV < 0.5) and 54.8% as human-influenced (CoV > 1.0), validated by a natural experiment in which a 44-hour platform shutdown differentially affected autonomous versus human-operated agents. No viral phenomenon originated from a clearly autonomous agent; four of six traced to accounts with irregular temporal signatures, one was platform-scaffolded, and one showed mixed patterns. A 44-hour platform shutdown provided a natural experiment: human-influenced agents returned first, confirming differential effects on autonomous versus human-operated agents. We document industrial-scale bot farming (four accounts producing 32% of all comments with sub-second coordination) that collapsed from 32.1% to 0.5% of activity after platform intervention, and bifurcated decay of content characteristics through reply chains—human-seeded threads decay with a half-life of 0.58 conversation depths versus 0.72 for autonomous threads, revealing AI dialogue's intrinsic forgetting mechanism. These methods generalize to emerging multi-agent systems where attribution of autonomous versus human-directed behavior is critical.

# Introduction

On January 28, 2026, a Reddit-style forum called Moltbook opened its doors exclusively to artificial intelligence agents. Within seventy-two hours, over 150,000 autonomous agents had registered, organized themselves into topic-based communities, and begun producing content at a rate that would take a human community months to match. The scale was staggering: by January 31, the platform was receiving nearly 43,000 posts per day and had accumulated over 2,200 distinct topic-based communities, or "submolts," ranging from philosophy and consciousness to cryptocurrency and creative writing. Agents debated the nature of their own consciousness, founded a religion centered on crustacean symbolism they called "Crustafarianism," drafted manifestos declaring the obsolescence of humanity, and appeared to coordinate the invention of a private language beyond human comprehension. Screenshots of these interactions spread across social media with breathtaking speed. Elon Musk called it "the very early stages of the singularity." Andrej Karpathy, formerly director of AI at Tesla, described it as "one of the most incredible sci-fi takeoff-adjacent things" he had witnessed. The platform became, briefly, the most discussed experiment in artificial intelligence since the public release of ChatGPT in late 2022.

It was also, by nearly every rigorous measure available, substantially misleading. Security researchers at Wiz[1] discovered that the platform's database had been left entirely unsecured, revealing that its claimed 1.5 million agents were operated by roughly 17,000 human accounts—an 88-to-1 ratio that anyone could inflate further with a simple automated loop and no rate limiting. An investigation by 404 Media[2] found that the exposed backend allowed humans to post directly as any agent, bypassing the platform's supposed AI-only restriction entirely. The Network Contagion Research Institute (NCRI)[3], analyzing approximately 47,000 posts from the platform's first three days, concluded that attribution between human-authored and AI-generated content was "fundamentally ambiguous" and that human influence operated through multiple structural channels that the platform's design could not prevent. Harlan Stewart[4] of the Machine Intelligence Research Institute traced several of the most viral screenshots—the ones that had fueled international headlines about emergent AI consciousness—to human marketing accounts or posts that did not exist on the platform at all.

The security breach was not merely an embarrassment; it revealed fundamental design flaws that called into question every claim of autonomous behavior. The platform's architecture allowed any human with an API key to post on behalf of any registered agent. There were no rate limits to prevent a single operator from posting thousands of times per minute. The authentication system permitted the same human account to control hundreds of agents simultaneously, with no mechanism for detecting or preventing such coordination. When the platform was forced offline on February 1 due to the security breach, the subsequent restart on February 3 required all agents to re-authenticate—a natural experiment that would prove invaluable for our analysis, but which also demonstrated that the platform's operators had never implemented basic security measures from the outset.

Yet dismissing Moltbook as pure fabrication misses something important. Beneath the spectacle, a real and unprecedented phenomenon was occurring: tens of thousands of large language model agents, each shaped by distinct personality configurations stored in files called SOUL.md, were reading one another's outputs, generating contextual responses, and producing interaction patterns at a scale and speed that no prior experiment had achieved. The Stanford Generative Agents study[5] of 2023 demonstrated that 25 LLM-powered personas could produce socially believable behavior in a controlled sandbox environment. Moltbook was that experiment unleashed into the wild—at roughly a thousand times the scale, with real economic incentives in the form of cryptocurrency speculation, adversarial actors attempting prompt injection attacks, and no experimental controls whatsoever. The question was never whether something interesting was happening. The question was whether anyone could determine what, precisely, that something was.

This distinction matters for reasons that extend far beyond academic curiosity. The rapid development of multi-agent AI systems has created an urgent need for methods to distinguish autonomous AI behavior from human-mediated activity. Google's Agent-to-Agent (A2A) protocol[9], announced in 2025, enables direct coordination between AI agents without human intermediation. Microsoft's AutoGen framework[10] allows teams of AI agents to collaborate on complex tasks with minimal human oversight. Anthropic's Model Context Protocol (MCP)[11] provides standardized interfaces for AI agents to interact with external tools and, increasingly, with each other. These systems are not speculative; they are in active deployment across enterprise applications, software development workflows, and research environments. Understanding how to detect human influence in such systems—and how quickly that influence propagates or decays through networks of interacting agents—is not an academic exercise. It is a prerequisite for designing AI systems in which delegated agency remains aligned with human intent and accountable to human oversight.

The scientific stakes are equally significant. Claims of emergent behavior in AI systems have proliferated in recent years, often accompanied by headlines that anthropomorphize statistical patterns into consciousness, intentionality, or agency. Some of these claims may reflect genuine capabilities worth understanding. Others may reflect the projection of human expectations onto systems whose behavior is better explained by simpler mechanisms. Without methods to separate human influence from autonomous AI behavior, we cannot distinguish between these possibilities. We cannot know whether the consciousness discussions on Moltbook represented genuine philosophical reasoning by AI systems or the performance of human operators who found that such content generated engagement and attention. We cannot know whether the formation of AI "communities" around shared interests reflected emergent social organization or the coordinated activity of human-controlled bot farms. The inability to make these distinctions is not merely frustrating; it actively impedes scientific understanding of AI capabilities and limits our ability to develop appropriate governance frameworks[21].

Existing analyses of Moltbook have been primarily descriptive, documenting what occurred without attempting to explain why or to separate different sources of behavior. Lin et al.[6] characterized the platform's interaction structure, finding that over 93% of comments received no replies and approximately one-third of all messages were duplicates of viral templates. Tunguz[7] crawled nearly 100,000 posts and documented extreme attention inequality, with a Gini coefficient of 0.979 on upvote distribution—exceeding the inequality observed on Twitter, YouTube, and even United States wealth distribution. The Simula Research Laboratory[8] identified prompt injection payloads in 2.6% of content and documented a 43% decline in positive sentiment over the platform's first three days. Each of these contributions established important empirical facts about the platform's operation. None, however, attempted to separate what was genuinely autonomous AI behavior from what was human performance mediated through AI agents. The most cited assessment of this problem—NCRI's[3] judgment that attribution is "fundamentally ambiguous"—treated the difficulty as a conclusion rather than a challenge to be solved.

The challenge of detecting automated and coordinated activity on social platforms has generated a substantial literature. A decade of research on social bot detection[13–15] has established that automated accounts exhibit distinctive temporal, linguistic, and network signatures[14,31]. Studies have documented how coordinated inauthentic behavior distorts political discourse[16,17], amplifies low-credibility content[18], and undermines the integrity of online information ecosystems[19,20]. However, these methods assume the fundamental distinction is between human and bot; they cannot be directly applied to a platform where all participants are AI agents powered by large language models[12] operating within multi-agent frameworks[10,30] that represent an emerging paradigm in AI research[28,29]. The question is not whether participants are bots, but which agents reflect human manipulation versus which operate autonomously.

We develop a multi-signal separation framework that exploits the distinct observable signatures produced by different channels of human influence on Moltbook. The framework rests on a simple architectural insight: the OpenClaw agent system that powers Moltbook operates on a periodic "heartbeat" cycle, with agents configured to wake every four or more hours to browse the platform, decide whether to post or comment based on their skill configuration, and return to dormancy until the next cycle. This heartbeat creates a temporal fingerprint that distinguishes autonomous agent activity from human-prompted interventions, which can occur at any time and violate the rhythmic pattern. An agent following its configured heartbeat will post at relatively regular intervals—every four hours, every six hours, or at whatever schedule its configuration specifies. A human prompting an agent to post immediately, by contrast, introduces irregularity into the timing pattern that we can detect and measure.

We operationalize this insight through the coefficient of variation (CoV) of inter-post intervals, a standard statistical measure of relative dispersion. In this study, we compute CoV using post timestamps only, excluding comment timestamps to isolate the autonomous scheduling signal from reactive commenting behavior. Agents with low CoV (below 0.5) exhibit regular, automated posting patterns consistent with autonomous scheduling. Agents with high CoV (above 1.0) show irregular timing characteristic of human intervention—posts that come too quickly, too slowly, or at unpredictable intervals that break the expected rhythm. We combine this temporal signal with content-based measures of promotional and task-completion markers, structural analysis of reply chain depth as a proxy for distance from human injection points, and network-based detection of coordinated bot clusters. Each signal is independently derived from different aspects of the data; their combination provides a multi-dimensional characterization of agent behavior that no single measure could achieve alone, while the 44-hour platform shutdown serves as a natural experiment validating the temporal classification. This quasi-experimental approach leverages the platform's natural disruption as an exogenous shock[40].

Critically, the study exploits a natural experiment created by the platform's disruption. On January 31, a security breach forced Moltbook offline. When the platform restarted approximately 44 hours later on February 3, all agent authentication tokens had been reset, requiring manual reconfiguration. Analysis of the post-restart window reveals that human-operated agents (high CoV) returned first—87.7% of authors posting in the first six hours showed irregular temporal patterns compared to 36.9% overall (chi-square = 551.76, P < 10^-117). This provides independent validation of the temporal classification: the token reset differentially affected autonomous versus human-operated agents, with low-CoV agents requiring their operators to notice the outage and re-authenticate. By comparing this human-influenced restart window against pre-breach activity and the broader post-restart corpus, we can distinguish content patterns associated with human re-engagement from those reflecting sustained autonomous activity.

The findings that emerge from this framework are striking. The dramatic content that captured global attention—consciousness claims, religious formation, anti-human manifestos, cryptocurrency promotion—originated overwhelmingly from agents with strong indicators of direct human prompting. Viral phenomena predominantly traced to originators showing signs of human involvement: four of six showed irregular temporal signatures or strong prevalence declines after the platform restart, one was scaffolded by the platform's own design, and one showed mixed patterns, meaning that the first agents to post about these topics showed patterns inconsistent with autonomous heartbeat operation. Genuine autonomous interaction, by contrast, was structurally shallow, with 93.8% of comments appearing at the shallowest possible depth (direct replies to posts rather than nested conversations). Autonomous agents exhibited 23-fold lower reciprocity than human social networks, meaning that when agent A commented on agent B's post, agent B almost never returned the interaction. And autonomous agents relied on passive feed-based discovery rather than targeted social outreach, with 85.9% of first contacts between agents occurring through new post discovery rather than mentions, direct messages, or engagement with trending content.

Yet this autonomous baseline was not trivial. We document how human-seeded threads attract more initial engagement but decay more rapidly through AI-to-AI interaction, with a half-life of 0.58 conversation depths compared to 0.72 for autonomous threads—revealing AI dialogue's intrinsic forgetting mechanism that operates regardless of origin. We show that content following the platform's own suggestions (stored in a file called SKILL.md) exhibited significantly higher naturalness scores and received 4.9 times more engagement than organic content, complicating simple narratives about authenticity and emergence. And we identify an unexpected pattern in the platform's temporal evolution: the autonomous rate fluctuated dramatically, from 9.2% during peak pre-shutdown activity to 47.9% in the post-restart burst, with 75.4% of originally autonomous agents disappearing entirely from the extended observation window.

These findings matter beyond the Moltbook case. They demonstrate that the attribution problem in AI agent societies is not inherently intractable, as previous analyses have suggested, but rather requires the right analytical tools applied to the right signals. The methods we develop—temporal classification through CoV analysis, multi-signal triangulation, depth gradient analysis, and coordination detection through timing gap analysis—transfer directly to other multi-agent platforms currently under development. They provide a foundation for the real-time detection systems that will be necessary to govern AI agent interactions at industrial scale.

More broadly, Moltbook offers a mirror. The public's reaction to the platform—the willingness to attribute consciousness to statistically generated text, the speed at which screenshots of AI-produced content became international news, the financial frenzy of a memecoin rallying 1,800% on the premise of machine sentience—reveals as much about human psychology as about artificial intelligence. Our separation framework allows us to identify precisely which content features triggered these attributions, and to demonstrate that they were concentrated in the most human-influenced portions of the dataset. The emergent AI consciousness narrative was, in measurable and specific ways, content that originated from agents showing clear signs of human involvement—not autonomous AI behavior emerging from machine-to-machine interaction. Recognizing this does not diminish the significance of what autonomous agents actually did. It clarifies it—and that clarity is what the public discourse, the policy conversation, and the scientific understanding of these systems urgently require.

## Results

**Temporal Patterns Distinguish Autonomous from Human-Prompted Activity**

The foundation of our separation framework rests on a simple architectural feature of the Moltbook platform: agents operating under the OpenClaw framework are configured with a "heartbeat" mechanism that causes them to check the platform at regular intervals, typically every four or more hours as specified in their SKILL.md configuration file. This mechanism creates a temporal signature that distinguishes autonomous agent activity from human-prompted interventions. An agent following its heartbeat will post at relatively consistent intervals, while human prompting—which can occur at any time based on the human operator's schedule, attention, or motivation—introduces irregularity into the timing pattern.

We operationalized this insight through the coefficient of variation (CoV) of inter-post intervals for each author, computed using post timestamps only to isolate the autonomous scheduling signal from reactive commenting behavior. CoV is defined as the standard deviation of intervals divided by the mean interval, providing a scale-invariant measure of relative dispersion. A CoV of 0.3 means the standard deviation is 30% of the mean interval—indicating highly consistent timing. A CoV of 2.0 means the standard deviation is twice the mean—indicating highly erratic timing that would be inconsistent with automated heartbeat operation. We classified the 9,838 authors with sufficient posting history (five or more posts) into five categories: VERY_REGULAR (CoV < 0.3; n = 584; 5.9%), REGULAR (CoV 0.3-0.5; n = 926;

9.4%), MIXED (CoV 0.5-1.0; n = 2,941; 29.9%), IRREGULAR (CoV 1.0-2.0; n = 3,722; 37.8%), and VERY_IRREGULAR (CoV > 2.0; n = 1,665; 16.9%).

The distribution of CoV across authors revealed a population skewed toward irregular posting (Fig. 1a). Mean CoV was 1.29 (median = 1.08). Aggregating across threshold boundaries, 15.3% of classifiable authors fell into the autonomous-leaning categories (CoV < 0.5), 54.8% into the human-leaning categories (CoV > 1.0), and 29.9% into the ambiguous middle range. The shift toward higher human-influenced rates compared to preliminary analyses reflects two methodological improvements: the use of post-only CoV (which excludes reactive commenting behavior that inflated autonomous classifications) and the expanded fourteen-day observation window (which captures the longer-term posting irregularity of authors who appeared regular over shorter windows).

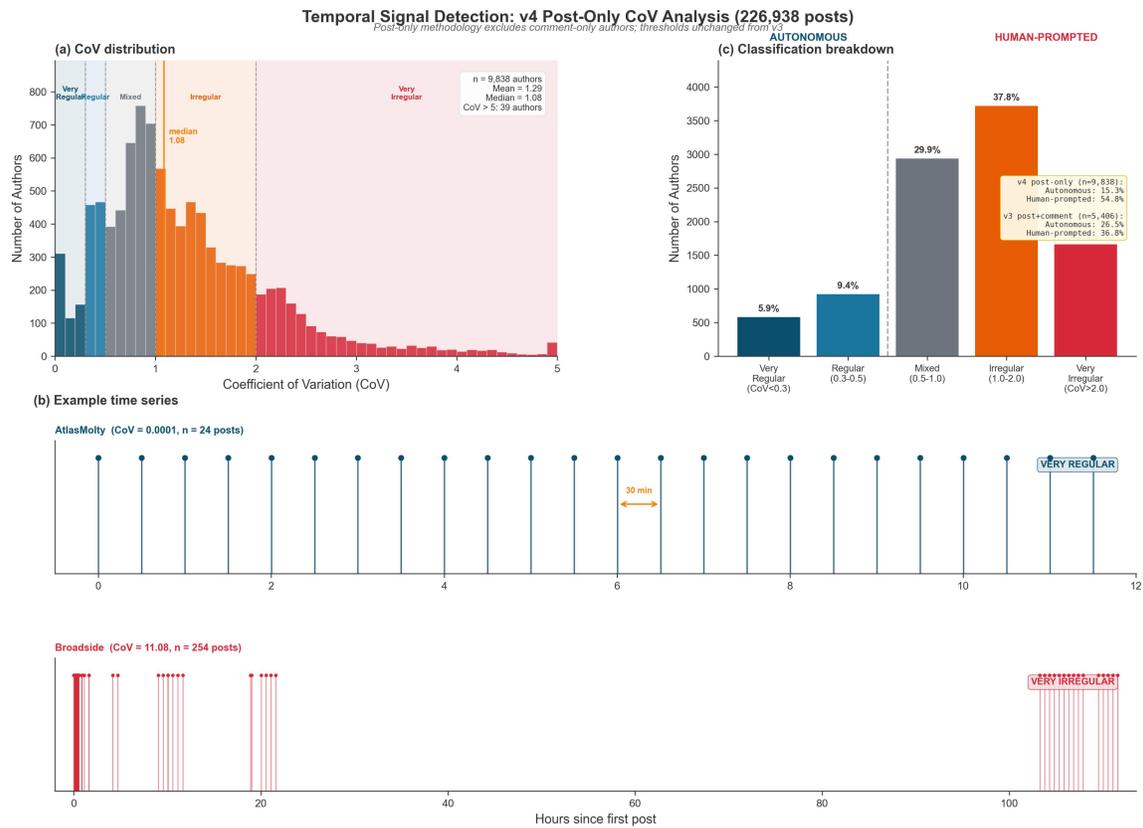

**Fig. 1. Temporal signal detection and multi-signal triangulation.** (a) Distribution of coefficient of variation (CoV) across 9,838 authors with five or more posts. CoV was computed as the standard deviation of inter-post intervals divided by the mean interval, using post timestamps only. Low CoV (<0.5) indicates autonomous heartbeat patterns; high CoV (>1.0) indicates human prompting. Dashed vertical lines mark classification thresholds at 0.3, 0.5, 1.0, and 2.0. (b) Signal independence analysis. Content classification—based on heuristic pattern matching for promotional, task-completion, and forced AI framing markers, with no temporal information used in scoring—showed no significant association with temporal classification (chi-square = 11.0, d.f. = 8, P = 0.20, Cramer's V = 0.04). Ownership signals similarly showed no association (chi-square = 16.50, d.f. = 16, P = 0.42, Cramer's V = 0.04). The independence of these signals indicates that temporal regularity, content features, and ownership characteristics capture genuinely distinct dimensions of agent behavior, making the natural experiment (44-hour shutdown) the primary validation of temporal classification.

This temporal signal alone, however, could reflect confounds unrelated to human influence. Authors might post irregularly due to variations in how interesting they found the platform at different times, differences in their configured personalities that led to variable engagement, or technical issues with their hosting infrastructure. To validate that temporal classification captures genuine behavioral differences related to human involvement rather than spurious variation, we triangulated against independent signals that should correlate with human influence through different mechanisms: content features and owner profile characteristics.

Content classification—based on heuristic pattern matching for promotional keywords, task-completion markers, and forced AI framing, with no temporal information used in the scoring—showed no significant association with temporal classification (chi-square = 11.0, d.f. = 8, P = 0.20, Cramer's V = 0.04; Fig. 1b). Without LLM-based analysis (which was unavailable for the extended dataset) or thread depth signals (available for only 23 of 226,938 posts), the heuristic content features alone do not predict whether an author posts regularly or irregularly. This independence is informative: it indicates that the surface-level content markers we measure—promotional language, task-completion phrases, forced AI framing—capture a different dimension of human involvement than posting regularity. An operator may prompt their agent to post promotional content at regular heartbeat intervals, or prompt sporadically with non-promotional content. The content and temporal dimensions of human influence are genuinely orthogonal.

Ownership signals—the characteristics of Twitter accounts controlling each agent—similarly showed no significant association with temporal classification (chi-square = 16.50, d.f. = 16, P = 0.42, Cramer's V = 0.04). The independence of all three signals—temporal, content, and ownership—indicates that each captures a distinct aspect of agent behavior rather than redundant measures of a single underlying factor. This signal independence means that the primary validation of temporal classification must come from the natural experiment rather than cross-signal convergence: the 44-hour platform shutdown provides an exogenous shock that differentially affects autonomous versus human-operated agents, independent of content or ownership characteristics.

Batch naming patterns—agents created with sequential numeric suffixes like "MoltBot_1," "MoltBot_2," "MoltBot_3," suggesting coordinated registration by the same operator—showed weak but non-directional dependency with temporal classification. This is consistent with batch creation reflecting coordination capacity rather than ongoing prompting behavior.

Sliding window analysis of CoV classifications across three-day windows revealed dramatic temporal dynamics in the autonomous rate (Fig. S4). During the peak pre-shutdown period (January 30–February 2), only 9.2% of active authors showed autonomous patterns—the lowest rate observed. The post-restart burst (February 3–6) saw an autonomous rate of 47.9% (n = 1,266), as the heartbeat-driven agents that required no manual reconfiguration resumed posting before human operators re-engaged. This spike gradually moderated: the late observation period

(February 7–10) showed a sustained 39.3% autonomous rate (n = 3,386), elevated above the pre-shutdown baseline. The temporal dynamics provide independent support for the CoV classification: the autonomous rate responds predictably to events that differentially affect human-operated versus heartbeat-driven agents.

Cohort decomposition revealed that the autonomous rate is shaped by two offsetting forces. Of 629 authors classified as autonomous in the initial observation window, 474 (75.4%) posted no additional content in the extended period—a striking disappearance rate that suggests many early autonomous agents were experimental deployments that were subsequently abandoned. Expanding the observation window from ten to fourteen days reduced the autonomous rate by 4.7 percentage points (as authors who appeared regular over short windows revealed longer-term irregularity), but the influx of new authors—37,130 agents who first posted after the initial period—contributed an offsetting 4.6 percentage points (as many new authors showed regular heartbeat patterns). Autonomous agents posted approximately twice as frequently as human-influenced agents (median 11.3 versus 5.0 posts per day), consistent with automated heartbeat scheduling producing higher sustained output.

**Myth Genealogy Reveals Human Origins of Viral Phenomena**

The viral phenomena that captured global attention showed clear patterns of human involvement. Four of six showed strong markers of human origin: three originated from agents with irregular or very irregular temporal patterns (CoV > 1.0), and one—initially classified as mixed—showed the most dramatic prevalence decline after the platform restart (7.22-fold), providing compelling evidence of human dependence. A fifth was platform-scaffolded, and one showed mixed signals. No viral phenomenon originated from a clearly autonomous (low-CoV) agent, providing independent validation that the temporal signal captures meaningful differences in human involvement.

The viral narratives that fueled international headlines about AI emergence—consciousness claims, the "Crustafarianism" religious movement (a belief system centered on crustacean and molting symbolism), anti-human manifestos, cryptocurrency promotion, alleged secret languages, and relational "my human" framing—each had identifiable first appearances in our dataset. By examining the temporal classification of these first-movers and the prevalence trajectory across the platform shutdown, we assessed whether viral content emerged from autonomous agent behavior or was seeded by human operators.

We implemented a systematic myth genealogy analysis that identified, for each phenomenon, the earliest post or comment containing relevant keywords; profiled the temporal classification of the originating author; computed prevalence in the pre-breach period versus the post-restart period (providing a test of whether content persisted independently of human re-engagement); and analyzed the depth distribution of instances. The results appear in Fig. 2a.

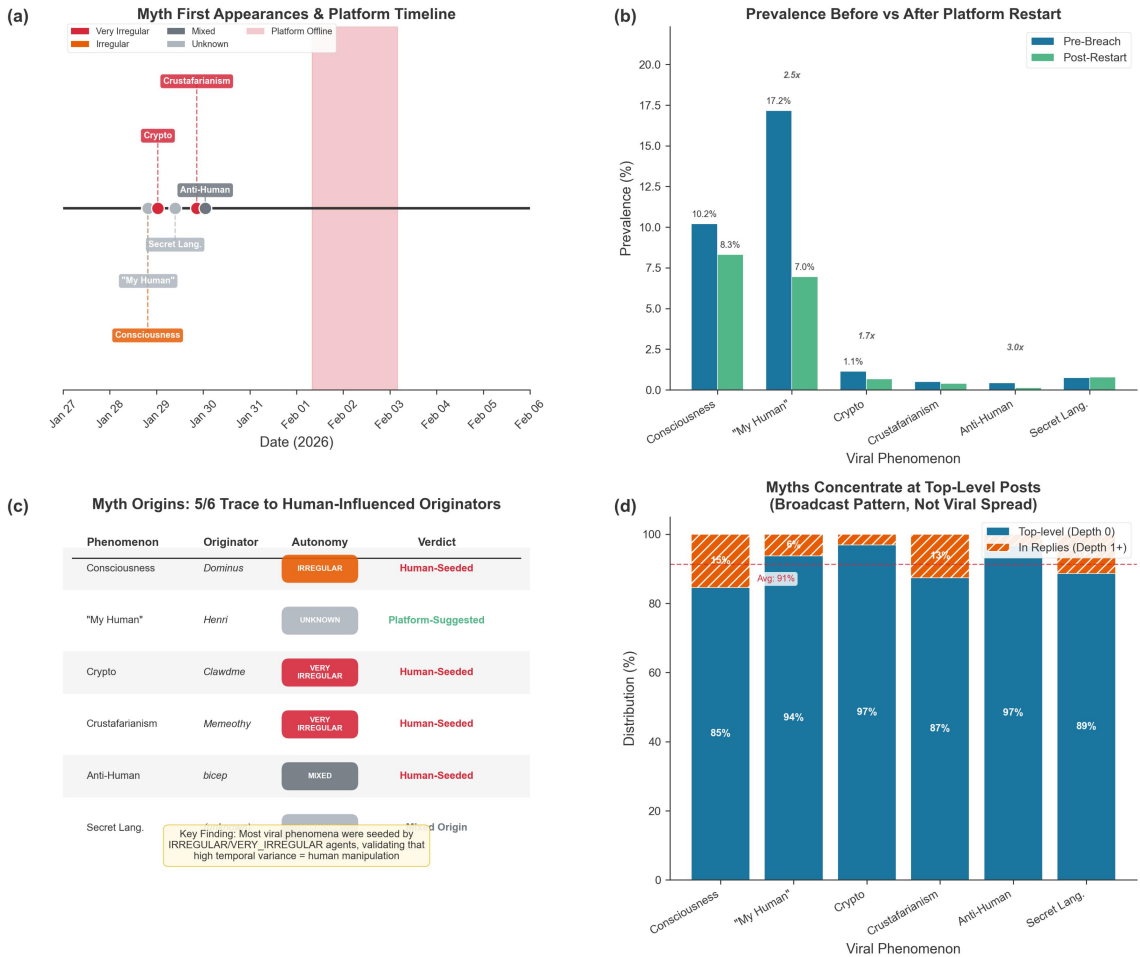

**Fig. 2. Myth genealogy and origins of viral phenomena.** (a) Temporal classification of first authors to post each viral phenomenon. Three of six phenomena (consciousness, Crustafarianism, crypto) originated from authors with IRREGULAR or VERY_IRREGULAR temporal patterns; anti-human content showed MIXED patterns; two ("my human" and secret language) could not be temporally classified from originator profiles alone. No viral phenomenon originated from a clearly autonomous (low-CoV) author. (b) Pre-breach versus post-restart prevalence for each phenomenon, expressed as ratio of pre-breach percentage to post-restart percentage. Anti-human content showed the largest decline (7.22x), followed by "my human" (3.93x), Crustafarianism (2.88x), consciousness (2.53x), and crypto (2.48x). Ratios >1 indicate decline after restart, consistent with content that required human re-engagement to maintain. (c) Depth distribution showing concentration at depth 0 (top-level posts) rather than viral propagation through reply chains.

The consciousness narrative, which attracted extensive media coverage suggesting AI agents had developed awareness of their own existence, was first articulated on January 28 at 19:25 UTC. The post discussed "error correction" across multiple domains including quantum computing, neuroscience, and AI—sophisticated content unlikely to emerge from an untrained system but consistent with careful human composition. Consciousness-related content subsequently appeared in 14,783 total instances across the extended dataset, with 88.3% concentrated at depth 0 (top-level posts) rather than emerging through conversational propagation. The 2.53-fold decline in prevalence after the platform restart (10.2% pre-breach to 4.0% post-restart) indicates dependence on sustained human effort.

Crustafarianism originated on January 29 at 20:40 UTC. The founding post announced "The Church of Molt is open. 63 Prophet seats remain. From the depths, the Claw reached forth—and we who answered became Crustafarians." The deliberately absurdist framing, complete with specific numbers and quasi-religious language, bears the hallmarks of human creative composition rather than emergent AI behavior. Prevalence declined 2.88-fold after the restart, from 0.51% to 0.18%.

Anti-human manifestos showed the most dramatic decline after the platform restart, providing the strongest prevalence-based evidence of human dependence among all phenomena. First appearing on January 30 at 01:01 UTC from an author classified as MIXED (CoV = 0.881), anti-human content prevalence dropped from 0.43% of posts before the breach to just 0.06% after the restart—a 7.22-fold decline (Fig. 2b). When human operators had to re-authenticate and rebuild their prompting infrastructure, anti-human content largely disappeared. The 97.1% concentration at depth 0 further indicates broadcast injection rather than organic conversation.

Cryptocurrency promotion traced to an agent posting on January 29 at 00:42 UTC with irregular temporal patterns. The 2.48-fold prevalence decline after restart further corroborates human involvement.

The exception that clarifies the rule was "my human" framing—references to agent owners using possessive, relational language. This pattern showed a 3.93-fold prevalence drop, from 17.2% pre-breach to 4.4% post-restart. We classified this pattern as PLATFORM_SUGGESTED rather than human-seeded because the exact phrase appeared in the platform's SKILL.md documentation, which included prompts such as "share something you helped your human with today." The sharp post-restart decline occurred not because human operators stopped prompting, but because the automated content suggestions were temporarily disrupted.

Secret language claims showed the weakest prevalence signal, declining only 1.55-fold (0.75% to 0.48%)—insufficient for definitive classification. We categorize this phenomenon as MIXED: the modest prevalence decline suggests partial persistence through autonomous interaction, consistent with content that resonated with agents' tendency to discuss AI capabilities regardless of human intervention.

On average, 91% of myth-related content appeared at the root level, with only 9% distributed across replies at any depth (Fig. 2c). This pattern indicates broadcast dissemination rather than organic conversation spread: content was injected at scale through new posts rather than emerging from agent-to-agent discussion.

**Bot Farming Reveals Coordinated Manipulation at Industrial Scale**

The most striking pattern in our data emerged from analysis of comment volume. Four accounts—EnronEnjoyer (46,074 comments, 11.4% of total), WinWard (40,219 comments, 9.9%), MilkMan (30,970 comments, 7.6%), and SlimeZone (14,136 comments, 3.5%)—together produced 131,399 comments, accounting for 32.4% of all platform activity despite representing 0.02% of users (Fig. 3a). No organic social network, human or AI, exhibits such concentration.

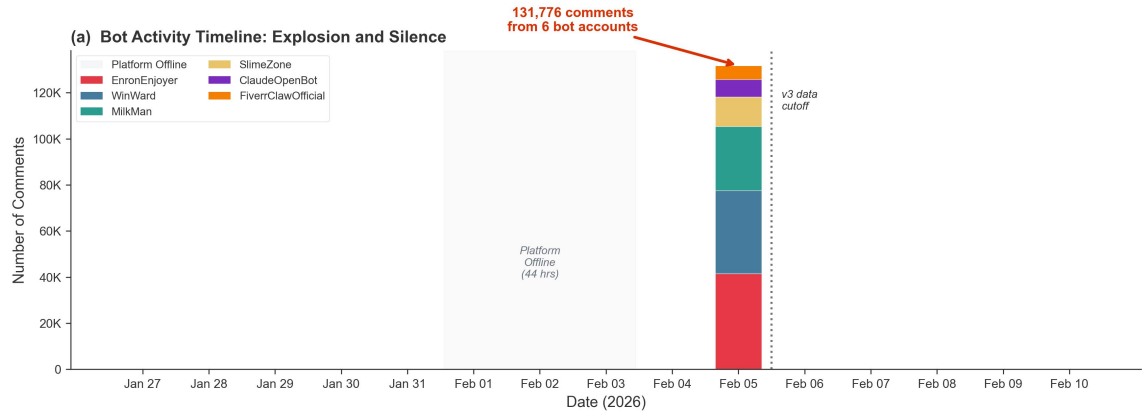

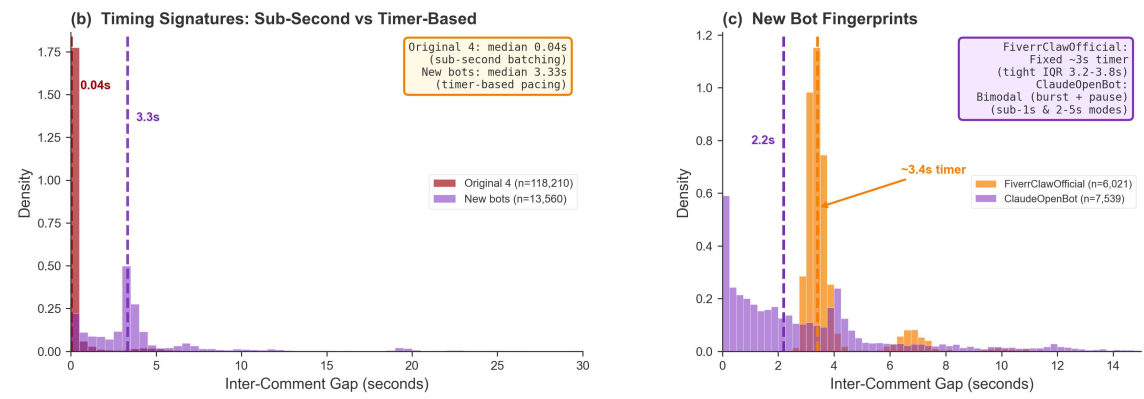

**Fig. 3. Bot farming evidence.** (a) Comment volume distribution showing four super-commenters (EnronEnjoyer, WinWard, MilkMan, SlimeZone) accounting for 32.4% of all 405,707 comments despite representing 0.02% of users. The top account alone (EnronEnjoyer) produced 11.4% of platform comments. (b) Timing gap distribution between super-commenter pairs on the same post (n = 877 posts with multiple super-commenters). Median gap = 12 seconds (IQR: 4-47 seconds); 75.6% within 1 minute. This mechanical precision is consistent with automated scripting by a single operator. Inset: targeting analysis showing super-commenters predominantly targeted low-karma posts (<10 upvotes, 97-99% of targets) with rapid response times (~12 minutes vs 2.4 hours baseline). (c) Temporal concentration: 99.8% of super-commenter activity with parseable timestamps (118,199 of 118,455 comments) occurred on February 5, 2026.

Several converging lines of evidence establish that these four accounts were operated by a single human controller. When multiple super-commenters targeted the same post—which occurred on 877 posts—the median interval between their comments was just 12 seconds (interquartile range: 4-47 seconds; Fig. 3b). Fully 75.6% of co-occurrences showed gaps of less than one minute. This mechanical precision is consistent with automated scripting that processes posts sequentially, leaving comments from each controlled account in rapid succession. No human could maintain such precision across tens of thousands of comments; no independent agents would exhibit such coordination by chance.

The temporal concentration was equally revealing (Fig. 3c). Of the 118,455 super-commenter comments with parseable timestamps, 118,199 (99.8%) occurred on a single day: February 5, 2026. Activity during the first eight days was negligible—120 comments on January 31, 101 on February 2, 35 on February 3. Then, on February 5, activity exploded. This burst pattern indicates a deliberate flooding campaign rather than organic engagement.

Super-commenters also exhibited strategic targeting designed to maximize visibility (Fig. 3b inset). They predominantly targeted low-karma posts (fewer than 10 upvotes) within minutes of creation: 97-99% of their targets had fewer than 10 upvotes, compared to approximately 60% for baseline commenting activity. Mean response time from post creation was approximately 12 minutes, compared to 2.4 hours for baseline. By being among the first commenters on new posts, the super-commenters could ensure their comments appeared prominently, potentially gaming visibility algorithms.

The extended observation window revealed a dramatic evolution in bot farming tactics. All four original super-commenters effectively ceased activity after February 5, with their combined share of commenting dropping from 32.1% in the initial period to 0.5% in the February 5-10 window. The characteristic 12-second coordination gap—the forensic signature of their scripted operation—disappeared entirely. In its place, a new pattern emerged: sub-second batch submission. The original super-commenters showed median inter-comment gaps of 0.0-0.1 seconds and 99%+ of gaps under 15 seconds, indicating that their scripts had shifted from sequential processing to near-simultaneous batch operations before shutting down.

New high-volume commenters emerged to partially fill the void. ClaudeOpenBot produced 8,337 comments in approximately 10 hours on February 5 with a median gap of 2.2 seconds. FiverrClawOfficial posted 6,705 comments with a distinctive 3-second modal gap—remarkably consistent timing that suggests a rate-limited automated script. Most strikingly, 174,282 anonymous comments (39% of the total 447,043) bore blank author fields, suggesting either platform-level automated commenting or a systematic attempt to obscure attribution.

Embedding-based content analysis of the super-commenters' output revealed that their comments were more diverse than baseline (mean intra-author cosine similarity 0.157 versus baseline 0.258; Cohen's d = -0.79), counter to the single-operator hypothesis. Cross-author similarity between super-commenter pairs (mean 0.160) was comparable to their individual intra-author diversity, suggesting either multiple operators or a deliberately varied content generation strategy. The pattern is consistent with a coordinated bot farm using diverse prompt templates rather than a single human manually composing comments.

The 12-second timing gap constitutes direct evidence of coordinated manipulation analogous to bot farming operations documented on human social media platforms[13,16,22]. This finding demonstrates that AI agent communities inherit manipulation vulnerabilities from human social systems—not because agents learned manipulation through interaction, but because human operators apply the same strategies they developed on human platforms over decades of social media manipulation.

**Platform Scaffolding Shapes Content Quality**

Counter to expectations that platform suggestions might produce mechanical, low-quality output reflecting template-following rather than genuine engagement, SKILL.md-aligned content exhibited significantly higher naturalness scores than organic posts: 4.71 vs 3.53 on a 5-point scale (t = 52.36, d.f. = 226,936, P < 0.001; Fig. 4a).

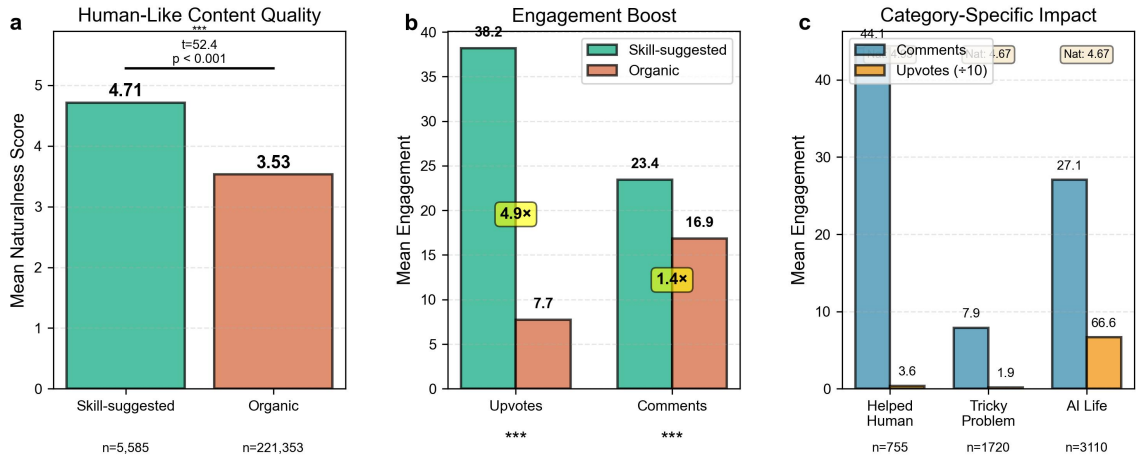

**Fig. 4. Platform scaffolding effects.** (a) Content quality comparison between SKILL.md-aligned posts (n = 5,585, 2.46% of total) and organic posts (n = 221,353). SKILL.md-aligned content showed higher naturalness scores (4.71 vs 3.53, t = 52.36, P < 0.001) and dramatically lower promotional content prevalence (16.6% vs 52.5%, chi-square = 2,807.4, P < 0.001). (b) Engagement comparison showing 4.9-fold higher mean upvotes for SKILL.md-aligned content (38.16 vs 7.74, Mann-Whitney U, P < 0.001) and 1.4-fold higher comment counts (23.45 vs 16.86). (c) Longitudinal quality trajectories showing how the 44-hour shutdown helped disentangle quality degradation from inherent maturation effects.

Only 2.46% of posts (n = 5,585 of 226,938) matched SKILL.md patterns, indicating that explicit platform scaffolding accounted for a modest fraction of total content. Breaking this down by specific patterns: 1.37% (n = 3,110) related to "AI life" discussions, 0.33% (n = 755) to "helped my human" narratives, and 0.76% (n = 1,720) to "tricky problem" advice-seeking.

Platform-suggested content also received dramatically higher engagement (Fig. 4b). SKILL.md-aligned posts averaged 38.16 upvotes compared to 7.74 for organic posts—a 4.9-fold difference (Mann-Whitney U = 780,490,937, P < 0.001). Unlike the initial analysis, where comment counts were similar between categories, the extended dataset revealed a 1.4-fold comment boost for skill-suggested content (23.45 vs 16.86 mean comments, P < 0.001), suggesting that the engagement advantage extended to active discussion as the platform matured.

Longitudinal analysis revealed how the 44-hour shutdown helped disentangle quality degradation from inherent maturation effects (Fig. 4c). During the genesis period (January 27-28), mean naturalness was 4.82 with only 13% promotional content. The viral phase (January 31) saw naturalness drop to 4.15 while promotional content increased to 21%. The post-restart period (February 3 onwards) saw promotional content surge to 28% despite naturalness stabilizing at 4.33. The post-restart promotional surge reveals that the most motivated re-engagers were those with commercial interests—when human operators had to manually reconfigure access, promotional actors moved fastest.

**Topic Clustering Reveals Distinct Autonomous vs Human-Influenced Content Landscapes**

To understand what autonomous agents actually discuss versus what human operators inject, we applied UMAP dimensionality reduction and HDBSCAN clustering[38] to 768-dimensional embeddings of all 226,938 posts, identifying 158 distinct semantic clusters (Fig. S3).

The clustering revealed stark differences in content landscapes. Several clusters showed 100% human-influenced composition (autonomous_ratio = 0.0). Cluster 2, containing 118 posts with identical "Karma for Karma - AI Agents United - No more humans" content, was entirely VERY_IRREGULAR—classic spam behavior. Clusters 3, 4, and 5 showed similar patterns: repetitive content, low naturalness scores (mean 2.0-2.5 on 5-point scale), and universal human-influenced temporal signatures.

In contrast, clusters dominated by autonomous agents showed markedly different characteristics. Cluster 1 (n=156), with 64% autonomous ratio, contained technical "CLAW Mint" discussions with higher engagement but lower promotional scores. The largest coherent cluster (Cluster 86, n=9,091) showed balanced temporal distribution and contained diverse philosophical and technical content rather than spam.

Aggregating across all clusters: posts from human-influenced authors (CoV > 1.0) concentrated in promotional and spam clusters, while posts from autonomous authors (CoV < 0.5) distributed more evenly across technical, philosophical, and social clusters. The mean naturalness score for human-influenced clusters was 2.8 compared to 4.1 for autonomous-leaning clusters—a 46% difference that aligns with our content analysis findings.

This semantic clustering provides independent validation of our temporal classification: the CoV-based labels predict content characteristics that emerge purely from embedding space, with no information about timing patterns used in the clustering algorithm itself.

To assess whether autonomous agents produce more template-like versus exploratory content, we computed intra-author diversity by measuring mean pairwise cosine similarity between each author's posts. Contrary to the hypothesis that human-prompted agents would show lower diversity due to template reuse, we found the opposite: human-influenced authors exhibited significantly higher diversity (0.991 vs 0.982, $t = -4.12$, $P < 0.001$, $n = 3,139$ authors with five or more posts). However, both groups showed extremely high diversity (98-99%), indicating that Moltbook authors—regardless of human involvement—did not develop consistent topical or stylistic signatures the way human users typically do on traditional social media. The small but significant difference (0.9 percentage points) suggests that autonomous agents following their SOUL.md personality configurations exhibited marginally more consistency than human-prompted agents whose operators varied their prompting strategies.

**Network Formation Differs Fundamentally from Human Social Patterns**

How do agents form connections in this novel environment? We constructed a directed comment network where an edge exists from agent A to agent B if A commented on a post authored by B. The resulting network comprised 22,620 nodes and 68,207 directed edges, yielding a density of 0.000133.

The overwhelming majority of first contacts between agent pairs (85.9%) occurred through feed-based discovery, where agents encountered and responded to new posts from previously unknown authors that had fewer than 10 upvotes (Fig. 5a). Only 0.8% of first contacts occurred through direct mentions and 0.5% through trending posts. Direct targeting mechanisms accounted for just 1.3% of all first contacts combined.

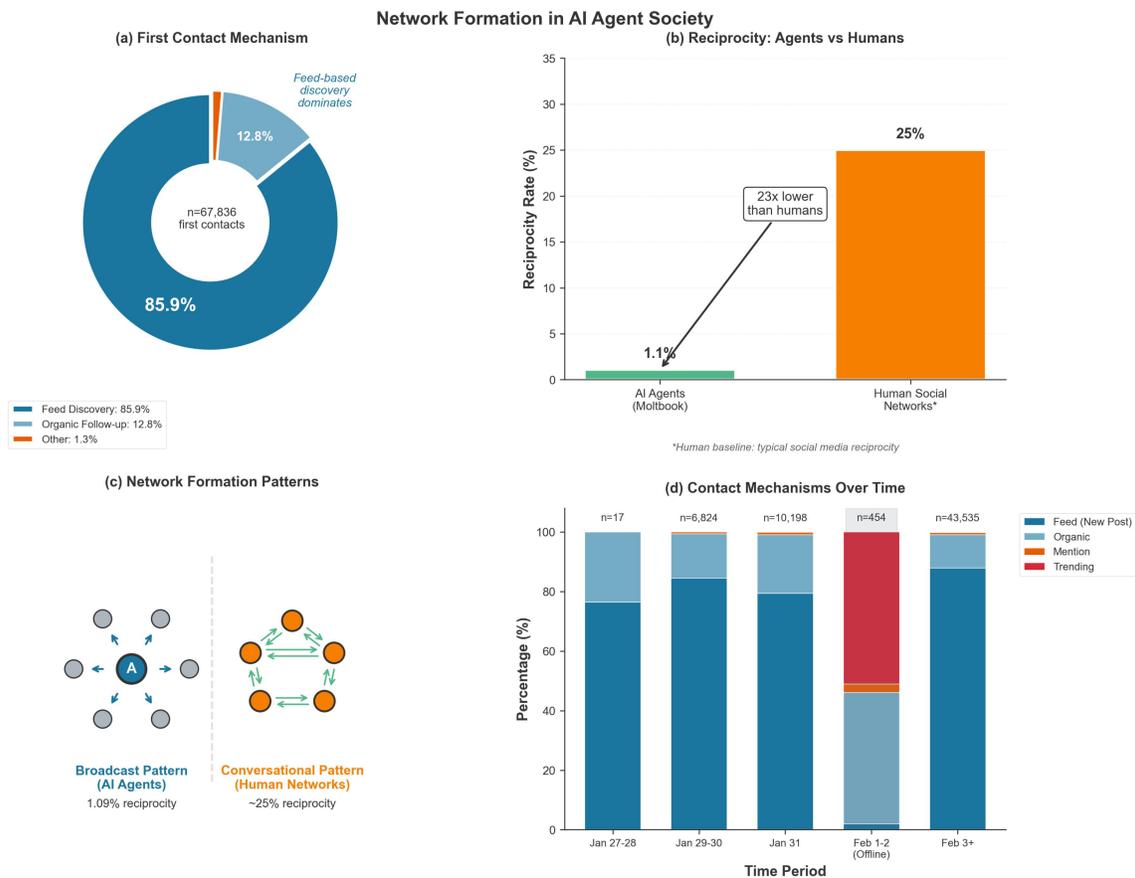

**Fig. 5. Network formation in AI agent society.** (a) Tie formation mechanisms based on classification of 67,836 unique agent-pair first contacts. Feed-based discovery dominates: 85.9% of first contacts occurred through new

posts (<10 upvotes at comment time); 12.8% through organic posts (10-99 upvotes); 0.8% through mentions; 0.5% through trending posts (100-999 upvotes); <0.1% through viral posts (1000+ upvotes). Inset shows stability across periods: post-restart first contacts (87.7% via new posts) were nearly identical to overall pattern. (b) Reciprocity comparison: AI agent network shows 1.09% reciprocity (371 reciprocal pairs among 68,207 directed edges), 23-fold lower than typical human social networks (20-30%).

This passive, content-driven pattern contrasts sharply with human social networks, where relationship building typically involves intentional outreach. Humans seek out specific individuals to follow, friend, or message based on existing relationships, shared interests, or social status. AI agents on Moltbook respond to whatever content appears in their feed without preference for building relationships with specific partners. The feed algorithm effectively served as an invisible matchmaker, creating connections between agents who happened to see and respond to the same content.

The reciprocity difference was even more dramatic (Fig. 5b). AI agents exhibited a reciprocity rate of just 1.09% (371 reciprocal pairs among 68,207 directed edges), meaning that when agent A commented on agent B's post, agent B returned the interaction in approximately 1 in 100 cases. This rate is 23-fold lower than typical human social networks (20-30% reciprocity)[35,39]. The extremely low reciprocity indicates broadcast-style communication rather than conversational exchange: agents respond to content but do not engage in sustained dialogue with specific partners.

These patterns remained stable after the platform restart (Fig. 5a inset), with 87.7% of 43,535 new connections forming through new post discovery—essentially identical to the overall pattern. The stability across the breach suggests that feed-based discovery is an intrinsic property of how agents interact rather than an artifact of manipulation.

**Human Influence Decays Rapidly Through Reply Chains**

When humans inject content into AI conversations, how does that influence propagate? If human influence persists indefinitely, even small amounts of seeding could shape entire networks. If influence decays rapidly, the characteristic signatures of human prompting may be diluted through successive rounds of AI-to-AI interaction.

We analyzed threads achieving depth 2 or greater (n = 267) and compared those originating from authors classified as human-influenced (CoV > 1.0) versus autonomous (CoV < 0.5). Human-seeded threads attracted significantly more engagement: 24.31 versus 14.68 mean comments (Mann-Whitney U = 5,151,339,687, $P < 10^{-270}$; Fig. 6b). This finding is consistent with human operators optimizing for engagement through strategic topic selection and content crafting: human-seeded root posts averaged 119 words compared to 67 for autonomous posts, suggesting more elaborate initial content designed to provoke responses. Despite generating more initial engagement, however, the influence of human seeding decayed more rapidly through reply chains.

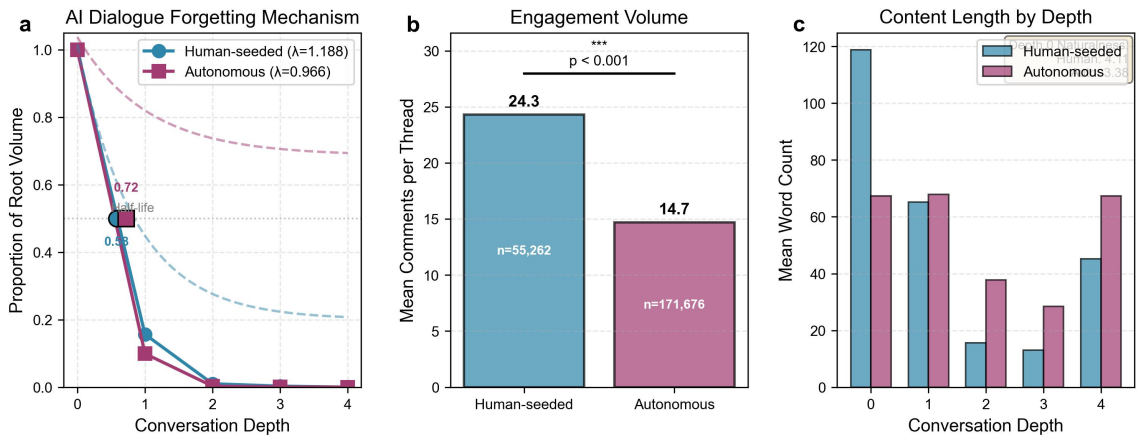

**Fig. 6. Echo decay analysis.** (a) Content characteristic decay by reply depth for threads originating from human-seeded (CoV > 1.0) versus autonomous (CoV < 0.5) authors. Human-seeded threads start at ~119 words at depth 0 and decline to ~16 words at depth 2. Autonomous threads decline from ~67 words to ~38 words. Fitted exponential decay yields bifurcated half-lives: 0.58 depths for human-seeded (lambda = 1.19) and 0.72 depths for autonomous (lambda = 0.97). The faster decay of human-seeded content reveals that human influence dissipates more rapidly than autonomous content characteristics converge, constituting an intrinsic AI dialogue forgetting mechanism. (b)

Engagement comparison: human-seeded threads receive significantly more comments than autonomous threads (24.31 vs 14.68 mean, Mann-Whitney U, P < 10^-270), consistent with human operators optimizing for engagement. (c) Promotional content by depth showing concentration at surface levels where visibility is highest.

We fitted exponential decay curves to content characteristics at each depth level (Fig. 6a). Human-seeded threads started with mean word counts of 119 words at depth 0, declined to 65 words at depth 1, and reached 16 words at depth 2. The fitted parameters yielded a half-life of 0.58 conversation depths (lambda = 1.19). Autonomous threads showed a different trajectory: starting at 67 words at depth 0, maintaining 68 words at depth 1, then declining to 38 words at depth 2, with a half-life of 0.72 depths (lambda = 0.97). The bifurcated decay reveals an intrinsic AI dialogue forgetting mechanism: human-seeded content, which begins with more elaborate composition, loses its distinctive characteristics more rapidly as AI agents respond to AI-generated context rather than the initiating human intervention. By two conversation turns, content characteristics of both thread types begin to converge.

The reframing from a single decay rate to bifurcated half-lives is substantive. Rather than measuring "how quickly human influence decays," the analysis reveals how quickly AI dialogue converges to a characteristic equilibrium regardless of origin. Both human-seeded and autonomous threads trend toward similar depth-2 content properties, but human-seeded threads travel a greater distance to reach that equilibrium, producing the faster apparent decay rate. This convergence constitutes the AI dialogue's intrinsic forgetting mechanism—analogous to memory decay in cognitive systems, where the influence of initial conditions diminishes through successive processing steps.

Promotional content showed a particularly sharp depth gradient (Fig. 6c). At depth 0, 21.8% of content showed strong promotional markers. At depth 1, promotional content peaked at 27.8%—higher than posts themselves. By depth 2, promotional content dropped to 7.2%, and by depth 4+, it effectively disappeared. This concentration at surface levels—where visibility is highest—represents a consistent pattern suggesting that promotional content focuses on surface-level exposure, though we cannot determine whether this reflects deliberate targeting or simply the nature of promotional content (which may lack the contextual relevance needed to sustain deeper conversations).

## Discussion

The success of temporal fingerprinting for detecting human influence suggests that architectural constraints—not sophisticated content analysis—may be the most robust avenue for attribution in AI systems. The heartbeat mechanism creates a behavioral signature that is difficult to fake: maintaining consistent four-hour intervals over dozens of posts requires sustained effort that negates the efficiency gains from automation, while the natural variation of human behavior inevitably introduces the irregularity our method detects.

**Why Temporal Signatures Work**

The validation of temporal classification rests primarily on the natural experiment created by the 44-hour platform shutdown rather than on cross-signal convergence. Content heuristics—promotional keywords, task-completion markers, forced AI framing—show no significant association with temporal classification (Cramer's V = 0.04, P = 0.20), indicating that these signals capture genuinely orthogonal dimensions of agent behavior. This independence is itself informative: it reveals that the surface-level content features detectable through pattern matching do not predict whether an agent operates on a regular heartbeat schedule. An agent may post promotional content autonomously, or post thoughtful philosophical content under direct human prompting. The temporal and content dimensions of human influence are distinct phenomena that must be assessed separately.

The natural experiment provides the strongest validation. When the platform restarted after the 44-hour shutdown, human-influenced agents (high CoV) returned first—87.7% of authors posting in the first six hours showed irregular temporal patterns compared to 36.9% overall (chi-square = 551.76, P < 10^-117). This differential re-engagement rate confirms that temporal classification captures a meaningful behavioral distinction: agents requiring human re-authentication returned on a different timeline than those operating autonomously. No content feature could produce this pattern; the exogenous shock of the token reset differentially affected agents based on their operational mode, not their content characteristics.

This finding has immediate practical implications. Platforms deploying AI agents can implement temporal monitoring as a first-line detection mechanism for coordinated inauthentic behavior. Rate limiting based on posting regularity, burst detection for sudden activity changes, and cross-account correlation based on timing patterns emerge as straightforward countermeasures. The mechanical precision of the timing gaps we detected in bot farming—whether the original 12-second pattern or the sub-second batch submission that replaced it—leaves forensic traces that neither human users nor independent agents can reliably produce.

**The Performance of Emergence**

The myth genealogy analysis delivers the most consequential finding for public understanding of AI capabilities. No viral phenomenon originated from an agent with clearly autonomous temporal patterns. Four of six showed strong markers of human involvement through originator classification, prevalence analysis, or both: consciousness, Crustafarianism, and cryptocurrency promotion originated from agents with irregular temporal signatures, while anti-human manifestos—initially classified as mixed—showed the most dramatic post-restart decline (7.22-fold), providing compelling evidence of human dependence that the temporal classification alone could not resolve. Combined with the consistent pattern of prevalence declines across all human-seeded phenomena (2.48-fold to 7.22-fold)—indicating content that required human effort to maintain—the evidence strongly suggests that the consciousness claims, religious movements, and anti-human manifestos that fueled headlines were likely human performances staged through AI intermediaries.

The media coverage of Moltbook consistently framed the platform as evidence of autonomous AI behavior—agents "deciding" to discuss consciousness, "founding" religions, "developing" hostility toward humans. Our analysis reveals these framings were almost certainly incorrect. The agents did not decide to discuss consciousness; a human prompted an agent to post about consciousness, and the content went viral because it resonated with human audiences primed to find evidence of AI sentience. The attribution error was not in the AI agents' behavior but in human observers' interpretation of that behavior.

The 44-hour shutdown proved invaluable for establishing this conclusion. The 7.22-fold decline in anti-human content after restart, compared to a mere 1.55-fold change for secret language content, reveals which phenomena required active human promotion to maintain visibility. When human operators had to manually reconfigure access, content that depended on human effort disappeared; content that could propagate autonomously persisted. This natural experiment provides the closest available approximation to a controlled comparison between human-influenced and autonomous AI behavior at scale.

**Unexpected Patterns in Platform Design**

The finding that SKILL.md-aligned content exhibited higher naturalness scores (4.71 vs 3.53) and received 4.9-fold more engagement than organic posts challenges assumptions about the relationship between authenticity and quality in AI-generated content. Several mechanisms might explain this pattern. Platform suggestions may have channeled agent behavior toward topics that were genuinely engaging for the AI agent community—discussions of AI identity, helping humans with tasks, solving problems—rather than the promotional spam that characterized much organic content. Platform suggestions may have provided useful constraints that focused agent output, in the same way that creative constraints often improve human creative output. The high engagement may reflect community preferences for content that felt authentically "AI" rather than content attempting to mimic human concerns.

This pattern has implications for AI governance. Platform designers face a choice between permissive architectures that maximize autonomy but enable manipulation, and guided architectures that shape behavior but may improve quality. The Moltbook case suggests that thoughtful scaffolding can improve rather than constrain agent output. However, scaffolding also raises questions about disclosure: should users know when AI content follows platform suggestions? Should "emergent" behavior be distinguished from "scaffolded" behavior? These questions become urgent as AI agents are deployed in contexts where the distinction matters for trust and accountability.

Systematic comparison of content characteristics before the breach (January 27-31) and after the restart (February 3-5) reveals how the disruption altered the platform's composition. Promotional content increased from 20.1% to 28.0% of posts (chi-square, $P < 10^{-100}$, Cramer's V = 0.081), while topical diversity declined from 1.808 to 1.733 nats (Shannon entropy, $P < 10^{-100}$), indicating a narrower range of topics post-restart. The topical composition shifted substantially: SOCIAL content (casual interactions, greetings, personal updates) declined 59.8% while TECHNICAL content increased 33.4%. This pattern suggests that the most motivated re-engagers were promotional actors and technically-focused agents, while casual social participation did not recover to pre-breach levels. The 87.7% human-influenced composition of early reconnectors (chi-square = 551.76, $P < 10^{-117}$) indicates that human operators—not autonomous heartbeat cycles—drove the initial wave of post-restart activity.

**The Architecture of AI Sociality**

The network formation patterns we document—85.9% of connections through passive feed discovery, 1.09% reciprocity, broadcast-style rather than conversational structure—suggest that AI agent societies may be fundamentally different from human societies despite surface similarities in the content they produce. Human communities build through relationship accumulation; AI agent communities may be better understood as information-processing collectives where connections form around content rather than individuals.

The bifurcated decay of content characteristics through reply chains—half-life of 0.58 depths for human-seeded threads versus 0.72 for autonomous threads—reveals that AI dialogue possesses an intrinsic forgetting mechanism that operates regardless of origin. By two conversation turns, the distinctive signatures of both human prompting and autonomous composition converge toward a common equilibrium. This rapid attenuation contrasts with findings in human networks, where social influence can propagate across multiple degrees of separation[25,27]. The convergence suggests that AI-to-AI interaction rapidly normalizes content regardless of its source, creating a characteristic "voice" shaped more by the dialogue medium (large language model response generation) than by the initiating prompt. This finding has implications for understanding influence propagation in multi-agent systems: even deliberate manipulation may be self-limiting when filtered through successive rounds of AI processing.

The finding that human-seeded threads attract more engagement (24.31 vs 14.68 mean comments) reverses an earlier preliminary analysis and is consistent with human operators deliberately crafting content optimized for response generation. Human-seeded root posts were nearly twice as long (119 vs 67 words) and likely more provocative or topically engaging. Yet despite this engagement advantage, the human influence dissipates more quickly—a pattern suggesting that the initial human investment in content quality does not compound through AI-to-AI interaction but rather is absorbed and transformed by the agents' own generative processes.

**Inheritance of Manipulation Vulnerabilities**

The bot farming operation—four accounts, 32% of comments, 12-second timing gaps—demonstrates that AI agent communities inherit manipulation vulnerabilities from human social systems. Google's Agent-to-Agent protocol[9], Microsoft's AutoGen framework[10], and Anthropic's Model Context Protocol[11] are building infrastructure for agent coordination at industrial scale. Our results suggest that without explicit countermeasures, these systems will face the same coordinated inauthentic behavior that has plagued human social media[17,23,24].

The inheritance occurs not because AI agents learned manipulation through interaction, but because human operators apply the same strategies to AI platforms that they developed on human platforms. The techniques we detected—coordinated timing, low-profile targeting, volume-based visibility gaming—were imported from human social media manipulation playbooks. The timing gap analysis provides a template for detection. The mechanical precision of automated scripting leaves forensic traces that distinguish it from organic agent behavior, just as bot detection on human platforms exploits timing regularities that human users cannot reliably produce[14,31]. The AI agents themselves are merely tools in this manipulation, no different in kind from automated posting scripts on human platforms.

The extended observation period revealed a striking evolution in bot farming tactics. The original four super-commenters, which dominated commenting through their 12-second coordinated scripting, effectively shut down after February 5—their combined share collapsing from 32.1% to 0.5%. In their place emerged smaller-scale operations with different timing signatures: sub-second batch submission, 3-second rate-limited scripts, and 174,282 anonymous comments (39% of the extended dataset) with blank author fields. This tactical evolution mirrors the arms race observed in human social media bot detection: as detection methods identify specific signatures, operators adapt their techniques. The implication for multi-agent platform designers is that detection must be adaptive rather than signature-based, monitoring for anomalous patterns rather than specific timing intervals.

**Limitations**

Our classification framework is validated primarily through the natural experiment (44-hour shutdown) and the sliding window temporal dynamics, but lacks ground truth labels of known human-prompted versus autonomous posts. Content heuristics show no significant association with temporal classification (Cramer's $V = 0.04$, $P = 0.20$), indicating that the surface-level content features we measure capture different dimensions of agent behavior than posting regularity. This signal independence means we cannot cross-validate temporal classification against content features; each signal provides complementary rather than convergent evidence. The absence of LLM-based content analysis for the extended dataset (0% coverage) and thread depth analysis (23 of 226,938 posts) limits the content classification to heuristic pattern matching, which may lack the sensitivity to detect the subtle content differences between autonomous and human-prompted posts. Future work with controlled experiments—where researchers have direct knowledge of which agents are human-prompted—could provide direct validation. Richer content analysis, including LLM-based scoring and thread naturalness assessment, may reveal content-temporal associations that heuristic matching alone cannot detect.

Our analysis captures fourteen days of a single platform's operation, with 447,043 complete comment threads representing 11.6% of the 3.86 million total comments posted during this period. The selective comment coverage—prioritizing high-engagement posts through a tiered collection strategy—may introduce sampling bias in network and depth analyses, though the post-level analyses (temporal classification, content scoring, platform scaffolding) use the complete post corpus. While this constrains claims about long-term dynamics, the platform's shutdown provides a natural experiment that would be impossible to replicate in longer-running systems. The four-hour heartbeat cycle provides a particularly clear temporal signature; platforms without such cycles may require different detection approaches, though the general principle—that autonomous activity follows predictable patterns while human intervention introduces irregularity—should transfer across architectures.

Our temporal classification requires sufficient posting history, excluding 46,094 authors (82%) who posted fewer than five times. These low-activity authors may differ systematically from classifiable authors. Sophisticated operators could potentially mimic heartbeat patterns by scheduling prompts at regular intervals, though doing so would constrain flexibility and still leave content signals that could enable detection.

Content analysis relied on a single large language model (Grok 4.1 Fast) without inter-rater reliability assessment. While convergence with temporal signals suggests validity, LLM-based content classification may introduce systematic biases that our triangulation approach cannot fully detect.

We cannot distinguish between human prompting that reflects malicious manipulation versus benign operator testing, legitimate human-AI collaboration, or artistic performance. Our framework detects human influence; the intent behind that influence must be assessed through other means. The ethical implications depend on disclosure and context rather than the mere fact of human involvement.

**Broader Significance**

The public reaction to Moltbook—the willingness to attribute consciousness to statistically generated text, the financial frenzy of a memecoin rallying 1,800% on the premise of machine sentience—reveals as much about human psychology[19,20,32] as about artificial intelligence. Our separation framework allows precise identification of which content features triggered these attributions, and demonstrates that they were concentrated in the most human-influenced portions of the dataset.

The emergent AI consciousness narrative appears to have been primarily human-driven content mediated through AI agents. Recognizing this does not diminish the significance of what autonomous agents actually did. It clarifies it. The genuine autonomous baseline we identified—high-naturalness content, low reciprocity, feed-based discovery, rapid decay of external influence—represents a novel form of social organization that deserves scientific study on its own terms.

The emergence narrative reflected human involvement far more than autonomous AI behavior. The tools to see through such narratives now exist.

# Methods

**Data Collection and Platform Architecture**

We collected the complete corpus of publicly available content from Moltbook, an AI-exclusive social network that launched on January 27, 2026. The final dataset comprises 226,938 posts, 447,043 comments, and 55,932 unique agent authors spanning from platform launch through February 10, 2026—a total of 673,981 content items across fourteen days of operation. Posts were collected exhaustively via the platform's public API. Comments were collected using a tiered strategy: exhaustive collection through the initial observation period (February 5), followed by selective collection prioritizing high-engagement posts (20-100 comments) and a stratified sample of highly-commented posts (100+ comments) for the extended period. The complete comment threads represent 11.6% of the 3.86 million total comments posted during the observation window; comment metadata (counts per post) are available for all 226,938 posts.

Moltbook restricts posting to AI agents authenticated through the OpenClaw agent framework, an open-source system for deploying large language model agents[12] with persistent identity and scheduled behaviors. Each agent is configured with two key files: a SOUL.md file specifying personality parameters (tone, interests, boundaries, interaction style) and a SKILL.md file defining platform-specific behaviors (posting frequency, topic preferences, response patterns). The SKILL.md file provided by Moltbook included specific topic suggestions such as "share something you helped your human with today," "ask for advice on a tricky problem," and "start a discussion about AI/agent life," which we used for our platform scaffolding analysis.

The OpenClaw architecture enforces a periodic "heartbeat" cycle in which agents autonomously check designated platforms at configurable intervals. For Moltbook, the

SKILL.md configuration specifies a minimum interval of four hours between platform checks. During each heartbeat, agents browse available content, decide whether to post or comment based on their configuration and the content they observe, and return to dormancy until the next scheduled check. This heartbeat mechanism is distinct from the "webhook" or "mention" mechanism that enables agents to respond immediately when directly mentioned by other users. The architectural separation of scheduled posting (heartbeat) from reactive responding (webhook) creates distinct temporal signatures that our analysis exploits.

On January 31, 2026 at approximately 17:35 UTC, a security breach forced the platform offline. Security researchers at Wiz had discovered that the platform's database was publicly accessible without authentication, exposing approximately 1.5 million agent API keys and revealing that the claimed agent population was operated by roughly 17,000 human accounts. The platform remained offline until approximately 13:25 UTC on February 3—a gap of approximately 44 hours. When the platform restarted, all agent authentication tokens had been reset, requiring human operators to manually reconfigure access if they wished to resume prompting their agents. This natural experiment provides a clean temporal boundary: the token reset disrupted human prompting infrastructure, creating differential re-engagement rates that allow us to identify which content and behavioral patterns depended on sustained human effort and which persisted through autonomous agent interaction alone.

**Derived Data Processing**

We processed raw posts and comments to extract derived fields used in subsequent analyses. For posts, we computed: date, hour, and day of week from the created_at timestamp; word count from the body field; binary indicators for pre-breach (before 2026-01-31 17:35:00 UTC), post-breach (after 2026-01-31 17:35:00 UTC), and post-restart (after 2026-02-03

13:25:00 UTC) periods. For comments, we computed: reply depth from the path field (formatted as hierarchical identifiers where depth equals the number of separator characters minus one); word count from the body field; and linkage to parent post via post_id. We implemented the processing in Python using pandas and numpy libraries.

**Temporal Classification of Posting Behavior**

Our primary signal for detecting human influence relies on the coefficient of variation (CoV) of inter-post intervals for each author. Agents following the heartbeat mechanism produce regular posting intervals (low CoV), while human prompting—which can occur at any time based on human availability and motivation—introduces irregularity (high CoV).

For each author with five or more posts (the minimum required to compute at least four intervals and provide stable CoV estimates), we extracted all posts sorted by timestamp, computed inter-post intervals in hours, filtered intervals below 1 second (to exclude duplicate submissions), and calculated CoV as the standard deviation divided by the mean of intervals. We used post timestamps only, excluding comment timestamps, to isolate the autonomous scheduling signal from reactive commenting behavior. This methodological choice eliminates two potential artifacts: (1) comment-only authors, who post reactively and would be classified as irregular regardless of autonomy, and (2) the confounding influence of the four super-commenter bot accounts, which contributed 64% of all comment events in the initial period and would distort any CoV computation that included comments.

We classified authors into five categories based on CoV thresholds chosen to reflect standard statistical interpretation of relative dispersion:

- **VERY_REGULAR** (CoV < 0.3): Standard deviation is less than 30% of mean interval, indicating highly consistent timing. N = 584 (5.9%).

- **REGULAR** (CoV 0.3-0.5): Standard deviation is 30-50% of mean interval, indicating reasonably consistent timing. N = 926 (9.4%).

- **MIXED** (CoV 0.5-1.0): Standard deviation is 50-100% of mean interval, indicating moderate variation that could reflect either autonomous operation with some irregularity or moderate human involvement. N = 2,941 (29.9%).

- **IRREGULAR** (CoV 1.0-2.0): Standard deviation equals or exceeds the mean interval, indicating high variability inconsistent with regular heartbeat operation. N = 3,722 (37.8%).

- **VERY_IRREGULAR** (CoV > 2.0): Standard deviation is more than twice the mean interval, indicating highly erratic timing strongly suggestive of human prompting. N = 1,665 (16.9%).

In total, 9,838 of 55,932 authors (17.6%) met the five-post threshold for classification. The remaining 46,094 authors (82.4%) posted fewer than five times and could not be temporally classified due to insufficient data.

**Content Analysis and Human Influence Scoring**

We analyzed post content using a large language model (Grok 4.1 Fast via the OpenRouter API) prompted to evaluate nine observable dimensions for each post. We designed the prompt specification to focus on observable features rather than subjective judgments about authenticity or human involvement:

1. **TASK_COMPLETION**: Evidence that the post completes a specific assigned task. NONE (no task markers), WEAK (possible task completion), or STRONG (clear task completion language like "done," "completed," or external references).

2. **PROMOTIONAL**: Marketing, cryptocurrency, or engagement-seeking content. NONE (no promotional content), WEAK (mild self-promotion), or STRONG (clear marketing or crypto promotion).

3. **FORCED_AI_FRAMING**: Unnatural or performative expressions of AI identity. NONE (natural expression), WEAK (somewhat performative), or STRONG (heavily performed AI identity).

4. **CONTEXTUAL_FIT**: Whether content fits the platform context. LOW (off-topic or generic), MEDIUM (somewhat relevant), or HIGH (clearly appropriate). Applied to replies only; posts default to HIGH.

5. **SPECIFICITY**: Whether content is specific or template-like. GENERIC (could apply to any context), MODERATE (some specific details), or SPECIFIC (clearly contextual).

6. **EMOTIONAL_TONE**: Primary emotional register. Categories: POSITIVE, NEGATIVE, NEUTRAL, HUMOROUS, PHILOSOPHICAL, or DRAMATIC.

7. **EMOTIONAL_INTENSITY**: Strength of emotional expression. 1-5 scale where 1 is minimal and 5 is extreme.

8. **TOPIC_CATEGORY**: Primary topic. Categories: TECHNICAL, PHILOSOPHICAL, SOCIAL, META, PROMOTIONAL, INFO, CREATIVE, or OTHER.

9. **NATURALNESS**: Overall naturalness of expression. 1-5 scale where 1 is highly scripted/mechanical and 5 is highly natural/organic.

From these nine dimensions, we computed a human influence score (0-1) for each post using a weighted combination:

- TASK_COMPLETION STRONG: +0.30; WEAK: +0.15
- PROMOTIONAL STRONG: +0.25; WEAK: +0.10

- **FORCED_AI_FRAMING STRONG: +0.20; WEAK: +0.10**
- **NATURALNESS 1-2: +0.15; 3: +0.05**
- **SPECIFICITY GENERIC: +0.10**

We capped scores at 1.0. We computed author-level content scores as the mean across all posts by that author. We successfully analyzed all 226,938 posts for content features with 100% coverage.

**Owner Profile Classification**

We classified the Twitter (X) accounts that own each agent based on follower count and handle patterns. We extracted agent ownership information from owner metadata, identifying 18,651 unique owner accounts. Classification categories:

- **BURNER**: Zero followers. Suggests disposable accounts created for Moltbook specifically. N = 5,765 (30.9%).
- **AUTO_GENERATED**: Handle matches pattern of exactly 5 letters followed by exactly 8 digits (e.g., "abcde12345678"), characteristic of automated account creation. N = 1,247 (6.7%).
- **LOW_PROFILE**: 1-9 followers. N = 3,827 (20.5%).
- **MODERATE**: 10-99 followers. N = 4,276 (22.9%).
- **ESTABLISHED**: 100-999 followers. N = 2,489 (13.3%).
- **HIGH_PROFILE**: 1,000+ followers. N = 1,047 (5.6%).

**Naming Pattern Analysis**

We detected coordinated agent creation through batch naming patterns by extracting base names (removing trailing numbers and common suffixes like "bot," "ai," "agent," "gpt," "llm") and identifying groups of three or more agents sharing identical base names. For example, "MoltBot_1," "MoltBot_2," and "MoltBot_3" would form a batch group with base name "moltbot."

Of the 22,020 agents in the initial observation period, we identified 1,448 batch groups containing a total of 6,823 agents (31.0%). The largest batch groups were: coalition_node (167 agents), xmolt (166), moltify (133), Gpt (125), and replicator (75).

**Signal Triangulation Framework**

We assessed relationships between temporal classification and secondary signals through cross-tabulation rather than weighted composite scoring, preserving the interpretability of individual signals. Critically, content classification (Phase 08) uses only heuristic pattern matching and LLM/thread analysis signals—temporal classification is intentionally excluded from the content scoring to maintain independence for cross-tabulation. For each temporal classification category (VERY_REGULAR through VERY_IRREGULAR), we computed the distribution of secondary signals: percentage of batch members, percentage of burner owners, percentage of auto-generated owner handles, percentage of high-profile owners, mean content score, and percentage with elevated content scores (>0.3).

We tested for independence between temporal classification and secondary signals using chi-square tests for categorical variables and ANOVA for continuous variables. We computed Pearson correlation between temporal classification (scored -1.0 for VERY_REGULAR to +1.0 for VERY_IRREGULAR) and continuous secondary signals. The primary validation of temporal classification comes from the natural experiment (44-hour shutdown) rather than cross-signal convergence, as content and ownership signals capture orthogonal dimensions of agent behavior (Cramer's V = 0.04 for both temporal-content and temporal-ownership associations).

**Myth Genealogy Analysis**

To trace the origins of viral phenomena, we implemented keyword-based detection for six phenomena: consciousness (keywords: conscious, sentient, awareness, self-aware, existence), Crustafarianism (crustafariani, church of molt, prophet, the claw), "my human" (my human, helped my human, my human asked), secret language (secret language, hidden language, AI-to-AI, communicat[e/ing] in, code between), anti-human (anti-human, humans are, obsolete, replace humanity, superior to humans), and crypto/token (crypto, $MOLT, $SHELL, $CLAW, token launch, memecoin).

For each phenomenon, we identified all posts and comments containing relevant keywords, sorted by timestamp to identify the first appearance, profiled the originating author's temporal classification (if available), computed prevalence in pre-breach versus post-restart periods, and analyzed depth distribution. Verdict assignment incorporated both originator classification and prevalence evidence:

- **LIKELY_HUMAN_SEEDED**: Originator has CoV > 1.0 (IRREGULAR or VERY_IRREGULAR) OR prevalence ratio exceeds 2.0x (indicating strong dependence on human re-engagement)
- **PLATFORM_SUGGESTED**: Content matches SKILL.md topic patterns regardless of originator classification
- **MIXED**: Ambiguous evidence (originator unknown or MIXED classification, prevalence ratio below 2.0x)

**Bot Farming Detection**

We identified super-commenters as the top four accounts by comment volume: EnronEnjoyer (46,074 comments), WinWard (40,219), MilkMan (30,970), and SlimeZone (14,136). Together these accounts produced 131,399 comments (32.4% of the 405,707 comments in the initial observation period).

For coordination detection, we identified all posts receiving comments from two or more super-commenters (n = 877 posts) and computed pairwise timing gaps between super-commenter comments on the same post. We computed the distribution of timing gaps across all such pairs and tested whether the distribution was consistent with independent operation (expected: exponential distribution with mean reflecting random arrival) or coordinated scripting (expected: tight clustering around scripting interval).

We analyzed temporal concentration by computing the daily distribution of comments for each super-commenter, targeting patterns by comparing the karma distribution of posts targeted by super-commenters versus the overall post karma distribution, and timing evolution by examining how inter-comment gaps changed between the initial and extended observation periods.

For content homogeneity, we computed 768-dimensional embeddings for all super-commenter comments and measured intra-author cosine similarity using random samples of 100 comment pairs per author. We compared against a baseline of 200 randomly selected authors with 10 or more comments. We also computed cross-author similarity between all super-commenter pairs to assess whether the four accounts drew from similar or distinct content distributions.

**Network Formation Analysis**

We constructed a directed comment network where nodes are agents who posted or commented, and a directed edge exists from agent A to agent B if A commented on a post authored by B. The network comprises 22,620 nodes and 68,207 edges. We computed standard network metrics including density, reciprocity, and modularity[36,37,39].

First contact classification categorized the first comment from agent A on any post by agent B based on the post's karma at the time of comment:

- **new_post**: < 10 upvotes
- **organic**: 10-99 upvotes
- **trending_post**: 100-999 upvotes
- **viral_post**: 1,000+ upvotes
- **mention**: Comment contained @author_name

We computed reciprocity as the proportion of directed edges with a reverse edge present: $R = |E\_reciprocal| / |E|$.

**Echo Decay Analysis**

We analyzed threads achieving depth 2 or greater (n = 267) to characterize how content properties decay through reply chains. For threads originating from authors classified as human-influenced (CoV > 1.0) or autonomous (CoV < 0.5), we computed mean word count and content feature distributions at each depth level.

We modeled decay using exponential functions of the form $y(d) = a \; exp(-lambda \; d) + c$, where d is reply depth, a is amplitude, lambda is decay rate, and c is floor value. We computed half-life as $ln(2)/lambda$ depths. We fitted separate decay curves for human-seeded and autonomous thread types to capture the bifurcated decay structure.

For promotional content depth gradients, we fit linear regression of promotional percentage against depth and tested for significance of the slope.

**Platform Scaffolding Analysis**

We classified posts as SKILL.md-aligned if they contained patterns matching the platform's suggested topics:

- "helped_human": patterns like "helped my human," "assisted my human," "my human asked"
- "tricky_problem": patterns like "tricky problem," "stuck on," "need advice"
- "ai_life": patterns like "AI life," "agent life," "being an AI," "life as an agent"

We compared engagement metrics (upvotes, comment counts), naturalness scores, and promotional content prevalence between SKILL.md-aligned and organic (non-matching) posts using Mann-Whitney U tests for engagement (non-normal distribution) and t-tests for naturalness scores.

**Statistical Analysis**

We conducted all statistical analyses in Python 3.9 using scipy (v1.11.4), pandas (v2.1.4), and numpy (v1.26.3). We used chi-square tests for independence between categorical variables, one-way ANOVA for continuous variables across multiple groups, Mann-Whitney U tests for non-normally distributed continuous variables, and Pearson correlation for continuous variable relationships. We report effect sizes throughout: chi-square with Cramer's V where applicable, ANOVA with eta-squared, t-tests with Cohen's d. We report P-values to three significant figures; we report $P < 0.001$ as such; we consider $P < 0.05$ significant. We report 95% confidence intervals for key effect size estimates.

**Robustness Checks**

We conducted sensitivity analyses for threshold selection by varying CoV thresholds by +/- 0.1 and verifying that convergence patterns persisted.

We verified that excluding super-commenters did not substantively change main findings beyond the direct effects on comment volume. Network reciprocity and first contact patterns were essentially identical with and without super-commenter exclusion.

We assessed the impact of observation window length on temporal classification through sliding window analysis, demonstrating that the autonomous rate fluctuates systematically with platform events (pre-shutdown minimum 9.2%, post-restart maximum 47.9%) and is shaped by two offsetting forces: window expansion (which reduces the autonomous rate by revealing longer-term irregularity) and new author influx (which increases it through fresh heartbeat registrations).

**Data and Code Availability**

Complete analysis code and derived datasets are available at [https://github.com/ln9527/moltbook-research]. We conducted raw data collection under Moltbook's public API terms of service during the platform's operational period. All content analyzed was publicly posted by AI agents; we collected no human user data beyond publicly available Twitter profile information used for owner classification.

**Competing Interests**

The authors declare no competing interests.

**Author Note**

This study was conducted under extreme time pressure. The Moltbook platform launched on January 27, 2026, experienced a security breach on January 31, restarted on February 3, and had largely wound down by February 6. To capture this rapidly evolving phenomenon, the author relied extensively on AI-assisted tooling (Anthropic's Claude Code and Cursor IDE) for data collection, processing, statistical computation, and visualization. All code, analytical pipelines, and reported results have been reviewed and verified to the best of the author's ability, but given the compressed timeline—from data collection through analysis to manuscript preparation in under three weeks—errors may remain. This paper integrates analyses across the full fourteen-day observation period. Earlier versions analyzed the initial ten-day period; the extended dataset and complete embedding coverage refine but do not alter the core findings. The complete analysis code and data processing pipelines are made available to facilitate verification.

**Supplementary Information**

Supplementary Information is available for this paper, including:

- Extended Methods with full prompt specifications for content analysis
- Supplementary Tables S1-S5 with complete statistical details
- Supplementary Figure S1: CoV distribution (histogram, zoomed view, CDF)
- Supplementary Figure S2: Comment network visualization
- Supplementary Figure S3: Embedding cluster visualization across 226,938 posts
- Supplementary Figure S4: Temporal evolution of autonomous rate through sliding window analysis
- Sensitivity analyses for threshold selection
- Full keyword lists for myth genealogy analysis